\documentclass[letterpaper, 10 pt, conference]{ieeeconf} 

\IEEEoverridecommandlockouts               

\usepackage{float}
\usepackage{tablefootnote}
\usepackage{multicol}
\usepackage{multirow}
\usepackage{epsfig} 
\usepackage{times} 
\usepackage{amsmath} 
\usepackage{amssymb} 
\let\labelindent\relax
\usepackage[shortlabels]{enumitem}
\usepackage{caption}
\DeclareCaptionType{algo}[\textbf{Algorithm}][List of Examples]

\usepackage{algorithm,algpseudocode}
\usepackage{graphicx}
\usepackage{epstopdf}
\usepackage{booktabs} 
\usepackage{array} 
\newcolumntype{P}[1]{>{\centering\arraybackslash}p{#1}}
\newcolumntype{M}[1]{>{\centering\arraybackslash}m{#1}}
\newcolumntype{L}[1]{>{\raggedright\let\newline\\\arraybackslash\hspace{0pt}}m{#1}}
\newcolumntype{C}[1]{>{\centering\let\newline\\\arraybackslash\hspace{0pt}}m{#1}}
\newcolumntype{R}[1]{>{\raggedleft\let\newline\\\arraybackslash\hspace{0pt}}m{#1}}
\usepackage[T1]{fontenc}
\usepackage[svgnames,table]{xcolor}

\usepackage{enumitem} 

\usepackage{verbatim} 

\makeatletter
\let\NAT@parse\undefined
\makeatother

\usepackage{cite}
\usepackage{url}
\usepackage[draft]{hyperref}

\DeclareMathOperator*{\argmin}{arg\,min}

\newcommand{\ie}{\textit{i.e.}}
\newcommand{\eg}{\textit{e.g.}}

\newcommand{\hl}[1]{\textcolor{black}{#1}}

\usepackage{fancyhdr}
\title{LAVAPilot: \underline{L}ightweight U\underline{AV} Trajectory Planner with Situational \underline{A}wareness for Embedded Autonomy to Track and Locate Radio-tags}

\author{Hoa Van Nguyen, Fei Chen, Joshua Chesser, Hamid Rezatofighi and Damith Ranasinghe
	\thanks{Hoa Van Nguyen, Fei Chen, Joshua Chesser, Hamid Rezatofighi and Damith Ranasinghe are with the School of Computer Science,
		The University of Adelaide, SA 5005, Australia
		{\tt\small \{hoavan.nguyen,fei.chen,hamid.rezatofighi,josh
			ua.chesser,damith.ranasinghe@adelaide.edu.au\}
	}}
}

\begin{document}

	\maketitle
	\thispagestyle{fancy} 
	\pagestyle{empty}

	\begin{abstract}
		
		Tracking and locating radio-tagged wildlife is a labor-intensive and time-consuming task necessary in wildlife conservation. 
		In this article, we focus on the problem of achieving \textit{embedded autonomy} for a resource-limited aerial robot for the task capable of avoiding \textit{undesirable disturbances to wildlife}. We employ a lightweight sensor system 
		capable of simultaneous (noisy) measurements of radio signal strength information from multiple tags for estimating object locations.
		We formulate a new \textit{lightweight} task-based trajectory planning method---\textit{LAVAPilot}---with a greedy evaluation strategy and a void functional formulation to achieve situational awareness to maintain a safe distance from objects of interest. Conceptually, we embed our intuition of moving closer to reduce the uncertainty of measurements into LAVAPilot instead of employing a computationally intensive information gain based planning strategy. We employ LAVAPilot and the sensor to build a lightweight aerial robot platform with fully embedded autonomy for jointly tracking and planning to track and locate multiple VHF radio collar tags used by conservation biologists. Using extensive Monte Carlo simulation-based experiments, implementations on a single board compute module, and field experiments using an aerial robot platform with multiple VHF radio collar tags, we evaluate our joint planning and tracking algorithms. Further, we compare our method with other information-based planning methods with and without situational awareness to demonstrate the effectiveness of our robot executing LAVAPilot. Our experiments demonstrate that LAVAPilot significantly reduces (by 98.5\%) the computational cost of planning to enable real-time planning decisions whilst achieving similar localization accuracy of objects compared to information gain based planning methods, albeit taking a slightly longer time to complete a mission. To support research in the field, and conservation biology, we also \textit{open source} the complete project. In particular, to the best of our knowledge, this is the first demonstration of a fully autonomous aerial robot system where trajectory planning and tracking to survey and locate \textit{multiple} radio-tagged objects are achieved onboard.
		
		

	\end{abstract}

	\section{Introduction}\label{sec:intro}
	
	Recent advancements in unmanned aerial vehicles (UAVs) or so-called drones capable of carrying different sensors such as thermal, chemical or visual sensors have enabled new opportunities in a range of applications from surveillance to search-and-rescue~\cite{Murphy2008,dames2017detecting}. Furthermore, falling unit cost prices have significantly reduced the entry barriers to employing commercial drones for solving real-world applications~\cite{chung2018a}. However, efficiently realizing these applications necessitates UAVs to perform tasks autonomously, but UAVs are faced with limited computational and power resources. 
	
	In this work, we are interested in developing an aerial robot system with \textit{embedded autonomy} for searching and localizing multiple radio-tagged objects of interest whilst considering the practical limitation of computational power onboard a UAV as well as the application desire to not disturb wildlife during the aerial robot's mission.
	This is an emerging application area to support conservation biologists to manage and protect endangered species and their habitats~\cite{wich2015drones,Shafer2019}. Over the last 50 years, VHF (Very High Frequency) tags can be considered as the most common technology for monitoring wildlife because of its versatility for different sizes and mass of animals and low cost~\cite{wikelski2007going}. 
	In contrast to vision-based methods {\cite{selby2011autonomous,zhou2013thermal,olivares2015towards,gonzalez2016unmanned}}, radio-collaring provides an effective method to identify and monitor individual animals of interesting species, especially significant in monitoring the well-being of endangered species re-introduced into their natural habitats ~\cite{kays2011tracking,thomas2012wildlife,tremblay2017low,webber2017radio}.
	Traditionally, conservation biologists have to trek through wildlife habitats using a handheld directional antenna to localize radio-tagged wildlife, which is a time-consuming and tedious task. 
	An autonomous UAV platform can: \textit{i)} contribute to automating the manual and tiresome task of localizing wildlife, often too difficult inaccessible terrains; \textit{ii)} reduce the costs involved in conducting field studies; \textit{iii)} provide \hl{a cost-effective} solution to gaining a height advantage to obtain better signal information with the ability of a UAV to fly above possible objects such as trees and shrubs intervening between a radio collar and a human receiver that can attenuate radio waves to ground-based receivers, and \textit{iv)} add the capability to gather more granular spatio-temporal information of endangered species via \textit{an autonomous} aerial system.

	\begin{figure} [!t]
		\centering
		\includegraphics[width=0.45\textwidth]{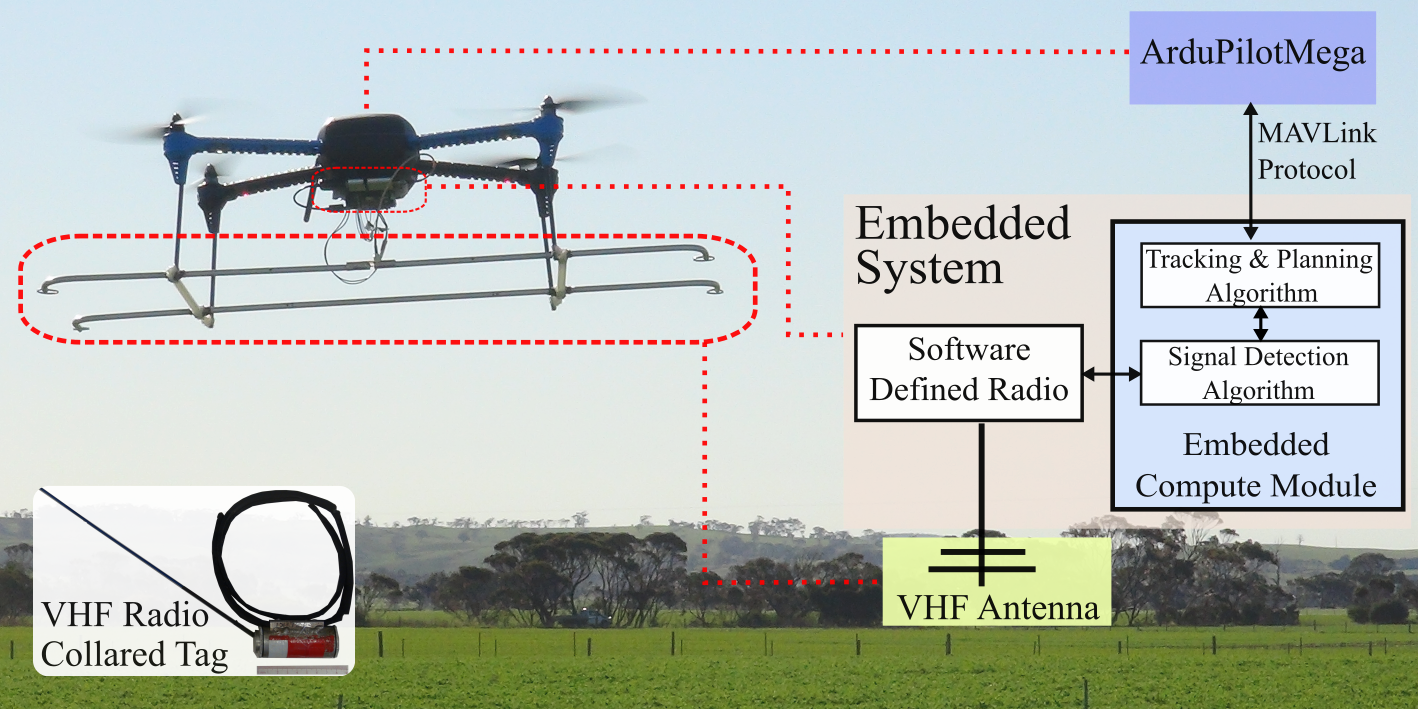}
		\caption{The autonomous UAV platform. Inset: VHF radio tag.
		}
		\vspace{-0.5cm}
		\label{fig_uav_system}
	\end{figure}

	We design an aerial robot system with \textit{embedded} autonomy for tracking and localizing \textit{multiple} radio-tagged objects. We formulate a real-time and online tracking and trajectory planning algorithm capable of execution onboard a resource-limited aerial robot platform; thus, eliminating the need for a cumbersome and battery-powered Ground Control System (GCS) set up in the field for the planning and tracking task. 
	We recognize that a closer approach by a UAV in a tracking task may disturb the wildlife of interest, as reported in recent studies~\cite{hodgson2016best,mulero2017unmanned}. \textit{Although it is sometimes desirable to agitate wildlife~\cite{paranjape2018robotic}, and disturbances can probably be  detrimental when trying to gather detailed spatial-temporal data of endangered species}. Therefore, we focus on formulating a trajectory planning algorithm to always maintain a safe distance from objects of interest. In summary, our main contributions are: 
	
	\vspace{2mm}
	\textbf{1)} We formulate a novel task-based trajectory planning method---we call \textit{LAVAPilot}---capable of being realized in an energy and computation limited platform. 
	We recognize that object position estimates are uncertain; however, object position uncertainty can be estimated from the belief density of the tracking algorithm. We also recognize that object signal measurement uncertainty is lower when the UAV is closer to the object. Therefore, our trajectory planning formulation is based on the intuition that the policy to navigate towards the objects with the lowest \textit{position estimation uncertainty} may provide a computationally efficient method of accurately tracking and localizing radio-tags---an important consideration for \textit{maximizing the number of objects located under the resource constraints of UAV platforms}.
	
	\vspace{2mm}
	\textbf{2)} 
	We incorporate a \textit{void probability functional} formulation to supply situational awareness to the trajectory planning algorithm to maintain a \textit{safe} distance from objects of interest during trajectory planning. We employ a \textit{greedy evaluation strategy} during the selection of optimal control actions to further reduce the computational complexity of planning under a void probability functional. We derive a lower bound probability that guarantees a UAV's capability to maintain a desired safe distance from all objects of interest.
	
	\vspace{2mm}
	\textbf{3)} We report the results of simulation studies and field experiments with multiple VHF radio-tags to evaluate our approach and demonstrate the capabilities of our aerial robot system. Our formulation leads to a computationally efficient planning method. To support research in the field, we also \textit{release the complete code base}\footnote{Complete system code release and project artifacts are available from \url{https://github.com/AdelaideAuto-IDLab/TrackerBots}.}.
	
	\vspace{2mm}
	In particular, to the best of our knowledge, this is the first demonstration of a \textit{fully} autonomous aerial robot system where trajectory planning and tracking to survey and locate \textit{multiple} radio-tagged objects are achieved onboard.
	\section{Related Work}\label{sec:related_work}
	An autonomous underwater system to localize radio-tagged carps was firstly proposed in \cite{tokekar2010robotic}. Subsequently, several underwater-based systems have been investigated~\cite{huntsberger2011intel,tokekar2013tracking,seto2013intro,vander2014cautious,hook2015algo,das2015data}. Early efforts using off-line estimation on multi-UAV to localize radio tags have been evaluated in \cite{frew2005radio,stachura2011cooperative,jensen2014monte}. Recently, the adoption of using software-defined radio (SDR) to detect multiple radio-tagged signals from a UAV has also been reported in \cite{dos2014small,vonehr2016software}. However, the problem of \textit{fully autonomous} tracking and localizing radio-tagged objects using UAVs is still open.
	
	Recently, two robotic aerial systems with a ground control station in the loop were demonstrated for localizing radio-tagged objects. The first study in~\cite{cliff2015online} located a ~\textit{single stationary} radio-tagged bird using bearing-only measurements from a phased antenna array. The second study employed received signal strength measurements to track and locate multiple mobile radio-tagged objects~\cite{hoa2017icra}. Bearing-only approaches lead to a bulky payload and substantially long measurement times (\eg, each bearing-only measurement takes $45$~s in  \cite{cliff2015online}). In contrast, range-only approaches are significantly faster, and using a lighter payload system~\cite{hoa2017icra}. The study in~\cite{hoa2017icra} employed an SDR to search and detect multiple radio-tagged signals to compute optimal control actions for a UAV using an information-based approach to track multiple mobile radio-tagged objects of interest. 
	However, both studies: \textit{i)} did not consider the problem of potential disturbances to animals in their path planning strategies, and \textit{ii)} relied on conservation biologists in the field to set up, power and employ a GCS in the loop to execute tracking and planning whilst employing an augmented communication channel with the drone platform for communications with the GCS---an added drain on the limited power capacity of the robot.

	Notably, a bearing measurement technique, similar to that in~\cite{cliff2015online}, as reported recently in \cite{bayram2017tracking}; however, their planning algorithm is calculated off-line using a sensor placement scheme for triangulation-based localization. Thus it is different from the approach in~\cite{cliff2015online} as well as our approach to developing a real-time and online joint tracking and path planning algorithm to develop a \textit{fully autonomous} system.
	
	Information gain-based planning approaches are proposed for tracking tasks in~\cite{cliff2015online,hoa2017icra,dames2017detecting,hoa2019tbd}. 
	However, information gain methods are demanding in both computations and power, often impractical under real-time requirements and/or demand a powerful ground control system in the loop for computations as in~\cite{cliff2015online,hoa2017icra}. In contrast, we are seeking to realize lightweight and real-time planning algorithms that can be embedded in the onboard low-cost drone platform.

	\section{Problem Formulation}\label{sec:problem}
	Our goal is to develop a fully autonomous system 
	and ensure trajectory planning decisions are supplied with situational awareness to reduce unwanted disturbances to radio-tagged objects of interest. In this section, we:~\textit{i)} describe the problem, \textit{ii)}~formulate the tracking algorithm, and \textit{iii)}~derive the trajectory planning algorithm with a void probability functional.

	\subsection{Problem Description}
	In our problem, biologists attach wildlife with VHF radio tags (as shown in the inset of Fig. \ref{fig_uav_system}) wherein the \textit{speed of a wildlife species is lower than the UAV's speed}. These radio collar tags transmit low-power on-off-keying signals every $T_0$ period and are detected by the sensor system on board a UAV. Therefore, we focus on tracking and localizing the detected objects instead of focusing on signal detection problems. In our problem, the UAV with state $u = [p^{(u)},\theta^{(u)}]^T \in \mathbb{R}^3 \times [0,2\pi)$, where $p^{(u)} = [p^{(u)}_{\mathrm{x}},p^{(u)}_{\mathrm{y}},p^{(u)}_{\mathrm{z}}]^T \in \mathbb{R}^3$ is the UAV position in 3D coordinate; $\theta^{(u)}$ is the UAV heading. The state of an object of interest is 
	$x = [p^{(x)}_{\mathrm{x}},p^{(x)}_{\mathrm{y}},p^{(x)}_{\mathrm{z}}]^T \in \mathbb{R}^3$, which is the object 3D position in $\mathrm{x,y}$ and $\mathrm{z}$ axes of the Cartesian coordinate system. 
	The total number of objects $N_{t}$ is assumed to be known; thus, the tracking and localizing operation task terminates when all the objects are tracked and localized. Further, to minimize disturbances to wildlife, the UAV must maintain a safe distance while navigating to track and localize wildlife. Although the UAV can fly at high altitudes (\eg, above $100$~m) to minimize disturbances to objects of interest, avoiding maneuvers above wildlife is also recommended ~\cite{mulero2017unmanned}. Further, flying at higher altitudes leads to significantly shorter flight times since it requires a considerable amount of onboard battery power to lift the UAV to high altitudes, especially for small UAVs. Therefore, in this work, we fix the UAV's altitude at $30$~m throughout the whole experiments to preserve the limited onboard battery power for the tracking and searching task and reduce computational demand for the embedded computer by not considering changing a UAV's altitude as a potential action when computing control actions.

	\subsection{Multiple Objects Tracking Formulation}
	In our problem, we do not consider missed detection and false detection; and objects of interest are uniquely identified by its frequency 
	estimated from measurements to provide a known association with objects. Therefore, our tracking algorithm is formulated using a sequential importance resampling particle filter (PF)~\cite{Gordon1993}, and an array of PFs is used to track multiple objects parallel. PF is a member of the Bayesian filters, which performs well in systems with nonlinear dynamic systems under noisy measurement data such as the received signal strength measurements affected by factors such as interference, radio-wave scattering and attenuation, and receiver thermal noise~\cite{daum2005nonlinear}. 
	
	Formally, let $x_k$ denote the object state at time $k$, which generates an observation $z_k$. Let $\pi_k(x_k|z_{1:k}) $ be the belief density of $x_k$ given the measurement history $z_{1:k}$. 
	The PFs use Sequential Monte Carlo (SMC) methods to approximate the belief density by a set of particles $\{\tilde{x}_k^{(i)}\}_{i=1}^{N_s}$ with its respective weights $\{w_k^{(i)}\}_{i=1}^{N_s}$,\ie,: 
	$
	\pi_k(x_k|z_{1:k}) \approx \sum\limits_{i=1}^{N_s} w_k^{(i)} \delta(x_k - \tilde{x}_k^{(i)} ), 
	$
	and $\sum\limits_{i=1}^{N_s} w_k^{(i)} = 1$, where $\delta(\cdot)$ is the Kronecker delta. 
	\vspace{2mm}

	\noindent\textbf{Measurement likelihood:} 
	The measurements in our problem are RSSI-based data related to radio-tags. Thus, we require a realistic signal propagation model to calculate the likelihood of a given measurement. In this work, we employ the \textit{MultiPath} model in~\cite{hoa2017icra}, a log distance path loss model with multi-path fading~\cite{wc1974microwave} to characterize the signal attenuation. In particular, the received power $h(x_k,u_k)$ [dBm] by the UAV at state $u_k$ from the transmitter of a radio-tagged object with state $x_k$ consists of the line-of-sight (LoS) transmitted power and the reflected power from the ground:
	\vspace{-0.2cm}
	\begin{align}
	h(x_k,u_k) &= P_0 - 10n \log_{10} (d(x_k,u_k)) \\ &+ G_r (x_k,u_k) + 10n \log_{10}(|1+\Gamma(\psi)e^{-j\triangle\phi}|). \notag
	\end{align}
	Here, $P_0$ is a constant which depends on the transmitted power; $n$ is a dimensionless path lost constant ranging from $2$ to $4$; $d(x_k,u_k)) = || x_k - p^{(u_k)} ||$ is the distance between the position of object $x_k$ and position of UAV $u_k$; $G_r(x_k,u_k)$ is a directional antenna gain depends on the relative position of object $x_k$ and UAV $u_k$, and the UAV heading angle $\theta^{(u_k)}$; $\Gamma(\psi)$ is the ground reflection coefficient dependent on the incidence angle $\psi$ between the ground plane and the reflected path; $\triangle \phi $ is the phase difference between the LOS path and the reflected path.
	
	The total measured power $z_k$ [dBm] is often corrupted by environmental noise $\eta^{(z)}$, given by:
	\vspace{-0.2cm}
	\begin{align}
	z_k = h(x_k,u_k) + \eta^{(z)}
	\end{align}
	
	Assuming that the object state dynamic process noise $\eta^{(x)}$ and the measurement noise $\eta^{(z)}$ are white noise, such that: $\eta^{(x)} \sim \mathcal{N}(0,Q^{(x)})$ and $\eta^{(z)} \sim \mathcal{N}(0,Q^{(z)})$, then the object state dynamics and the measurement likelihood function are:
	\begin{align}
	f_{k|k-1} (x_k|x_{k-1}) &= \mathcal{N}(x_k;f_{k-1}(x_{k-1}),Q^{(x)}) , \notag \\ 
	g_k(z_k|x_k) &= \mathcal{N}(z_k; h(x_k,u_k), Q^{(z)}),\notag
	\end{align}
	where $\mathcal{N}(\cdot;\mu,Q)$ denotes a Gaussian distribution with mean $\mu$ and covariance $Q$; and $f(\cdot)$ denotes a single-object transition function. 
	
	\subsection{LAVAPilot: Path Planning Algorithm Formulation}
	
	We are interested in formulating a real-time and online path planning algorithm that is suitable for computation and power limited embedded computing modules such as our aerial robot system in Fig. \ref{fig_uav_system}. In this work, we propose a new task-based planning method. We recognize that a UAV navigating towards an object where actual position estimates are unreliable can rapidly localize object positions. This is because the object signal measurement uncertainty is lower when the UAV is closer, an artifact of range-only tracking with signal strength measurements. Therefore: 
	
	\vspace{1mm}
	\noindent\textit{We hypothesize that in a tracking task with the objective of surveying objects to achieve an adequately small position estimation uncertainty, a planning strategy to move towards the object with the lowest position estimation uncertainty may provide an inexpensive planning strategy}. 
	
	Importantly, we have highlighted in Section~\ref{sec:intro}, the desire to reduce disturbances to wildlife in our tracking problem. Therefore, the planning algorithm needs awareness of radio-tags to maintain a safe ground distance between the UAV and the surveyed objects. Inspired by the void probability functional proposed in~\cite{beard2017void}, we consider planning informed by such a function to maintain a safe distance between the UAV and objects of interest. 
	Therefore, we formulate a task-based planning strategy based on \textit{selecting the control action that i) further minimizes the lowest estimated belief uncertainty of a given set of objects, and ii) maintains a safe distance between the UAV and objects of interest over a finite horizon time step $H$.} 
	
	In the following sections, we describe: \textit{i)} the formulation of the objective function for our task-based planning strategy, \textit{ii)} the formulation of the void functional to maintain a safe distance to reduce disturbances to objects of interest, and \textit{iii)} \hl{the selection of control actions}.

	\noindent\textbf{Objective Function Formulation:}
	We employ position estimation uncertainty represented by the filter belief density to formulate our objective function for evaluating control actions. Although using the maximum eigenvalue of the full covariance matrix of particles yields an E-optimal uncertainty estimation~\cite{pukelsheim2006optimal}, we adopt a computationally \textit{lower}-cost approach that measures the maximum covariance along with one of the coordinate axes with a similar performance guarantee. 
	
	Let $\hat{x} = \sum_{i=1}^{N_s} w_i \tilde{x}^{(i)}$ be the estimated object position from the particles set $\{w^{(i)},\tilde{x}^{(i)}\}_{i=1}^{N_s}$. Let $\sigma^{(x)}_x = \big[\sum_{i=1}^{N_s} w^{(i)} (p^{(\hat{x})}_x - p^{(\tilde{x}^{(i)})}_x)^2\big]^{1/2}$, $\sigma^{(x)}_y = \big[\sum_{i=1}^{N_s} w^{(i)} (p^{(\hat{x})}_y - p^{(\tilde{x}^{(i)})}_y)^2\big]^{1/2}$, and  $\sigma^{(x)}_z = \big[\sum_{i=1}^{N_s} w^{(i)} (p^{(\hat{x})}_z - p^{(\tilde{x}^{(i)})}_z)^2\big]^{1/2}$ represent the $x,y,z$ standard deviations of estimated belief density $\pi(\cdot)$. The standard deviation of the estimated belief density $\pi(\cdot)$ is defined as
	\vspace{-0.2cm}
	\begin{align} \label{eq:stdev_def}
	\sigma^{(x)} = \max(\sigma^{(x)}_x,\sigma^{(x)}_y,\sigma^{(x)}_z).
	\end{align}

	\noindent\textbf{Void Probability Functional Formulation}: 
	We formulate a void probability functional to maintain a safe distance $r_{\min}$ between the observer and objects of interest over a finite horizon $H$. Let $V(u, r_{\min})$ denote the void region of the object state based on UAV state $u$
	, given by:
	\begin{equation}\label{eq_void_region_def}
	\resizebox{.9\hsize}{!}{$V(u, r_{\min}) = \Big\{ x \in \mathbb{X}:\sqrt{(p^{(x)}_x - p^{(u)}_x)^2 + (p^{(x)}_y - p^{(u)}_y)^2} < r_{\min} $}
	\end{equation}
	
	
	Given a region $S \subseteq \mathbb{X}$, the estimated belief density $\pi_k(x_k|z_{1:k}) \approx \sum\limits_{i=1}^{N_s} w_k^{(i)} \delta(x_k - \tilde{x}_k^{(i)} )$; let $B_{\pi_k (x_k|\cdot)}(S)$ denote the void probability functional on $\pi_k (x_k|\cdot)$, then we obtain:
	\vspace{-0.2cm}
	\begin{align}\label{eq:B}
	B_{\pi_k(x_k|\cdot)}(S) = 1 - \sum_{i=1}^{N_s}w^{(i)} 1_S(\tilde{x}_k^{(i)})
	\end{align}
	where $1_S(\cdot)$ is the indicator function of region $S$ which is unity if $\tilde{x}_k^{(i)} \in S $ and zero otherwise. 

	\hl{Now we let $u_{k+1:k+H} = [u_{k+1},\dots, u_{k+H}]^T$ denote the trajectory of the UAV from time $k+1$ to $k+H$; $X_k = \{x_{1,k},\dots,x_{N_t,k}\}$ denote a set of objects at time $k$ where the belief density of each object $\pi_k(x_{j,k})$ is approximated by the particles set $\{w^{(i)}_{j,k},\tilde{x}^{(i)}_{j,k}\}_{i=1}^{N_s}$; the void probability functional of the trajectory $u_{k+1:k+H}$ is defined as:}
	\vspace{-0.2cm}
	\begin{align} \label{eq_void_prob}
	&B_{\pi_k}(V(u_{k+1:k+H},r_{\min})) \notag = \min_{\substack{ {x \in X_{k}} \\ {u \in u_{k+1:k+H}} }} B_{\pi_k(x|\cdot)}\big(V(u, r_{\min})\big) \notag \\
	&=\min_{\substack{ {j \in \{1,\dots,N_t \} } \\ {u \in u_{k+1:k+H}} }} \bigg[ 1-\sum_{i=1}^{N_s} w^{(i)}_{j,k} 1_{V(u, r_{\min})}(\tilde{x}^{(i)}_{j,k}) \bigg] 
	\end{align}

	\vspace{-0.2cm}
	\noindent\textbf{Control Action Selection:~} Given the formulation of the objective function and the void probability functional formulation, we now discuss our proposed planning algorithm employed to select the best control action policy.
	
	
	Our planning algorithm is summarized in Alg.~\ref{algo_guava}.
	Here, we apply the UAV motion model in \cite{hoa2017icra} to emulate a UAV control action resulting in trajectory $u_{k+1:k+H}$ over the look-ahead horizon time step $H$ while ensuring the UAV maintains a safe distance to objects of interest. We consider control actions for a quad-copter UAV, as shown in Fig.~\ref{fig_uav_system}. In practice, a quad-copter UAV can be controlled by providing way-points. For each controlled way-point $u_k^{c}$, we calculate the UAV trajectory over $u_{k+1:k+H}$ using a one-step Euler method~\cite{dames2017detecting} for $HT_0$ seconds with a very small integration time step (\eg, $1$~ms) to minimize the integration error; here $T_0$ (\eg, $1$~s) is the measurement duration. In the following, we describe our planning algorithm for selecting the optimal action over a look ahead horizon $H$.
	
\begin{figure} [!t]
		\centering
		\vspace{0.2cm}
		\includegraphics[width=0.23\textwidth]{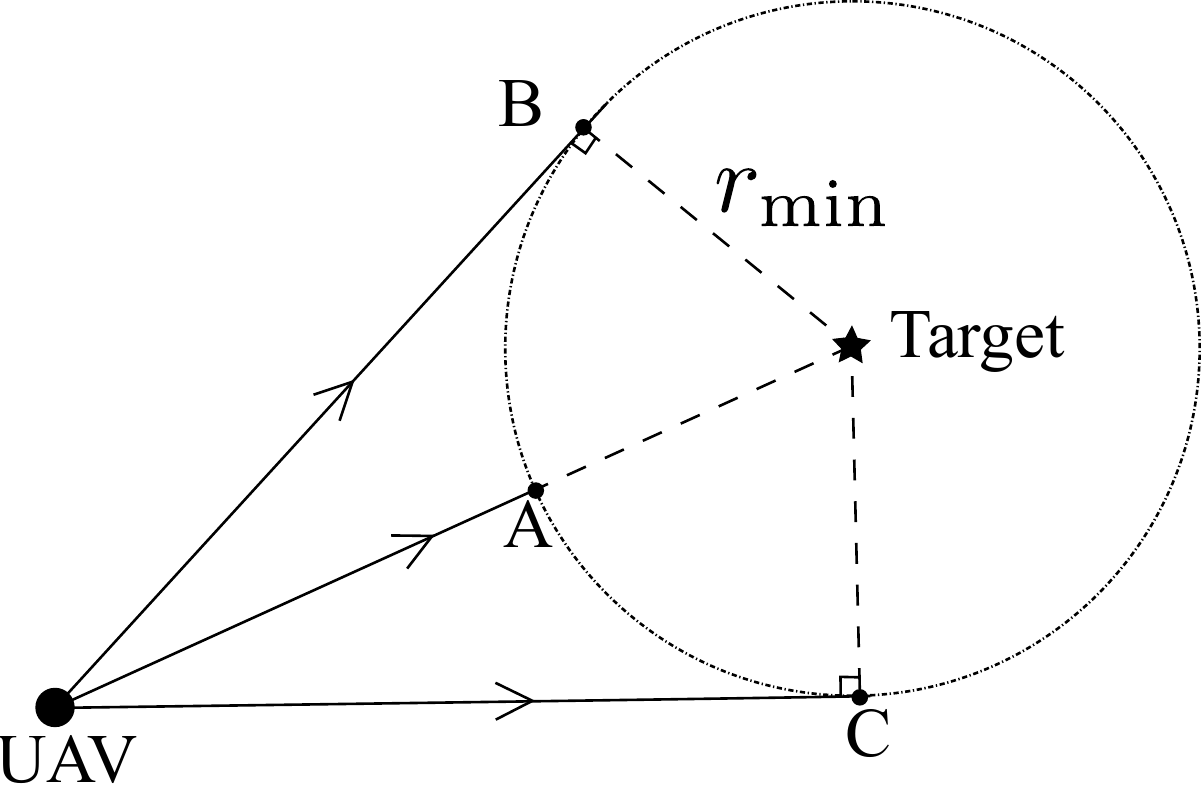}
		\vspace{-0.1cm}
		\caption{Illustration of the UAV trajectory planning actions \hl{at a fixed altitude} to rapidly reduce position estimation uncertainty under the void probability functional. Point A is the intersection point between the direct LOS path from the UAV to object; Point B and point C are the two tangential points on the circle defined by a void constraint.}
		\vspace{-0.4cm}
		\label{fig_Move_toward_an_object_with_void}
	\end{figure}

	Suppose $x_k^{*}$ is the object with the smallest estimation uncertainty that is not yet localized (\ie, $\sigma^{(x^{*}_k)} > \sigma_{\min}$). Fig. \ref{fig_Move_toward_an_object_with_void} illustrates a UAV with state $u_k$ with three possible control actions \hl{at a fixed altitude}\footnote{Although 3D tracking is feasible, as our initial investigation in~\cite{hoa2018acra}, we limit our investigation to a 2D domain to reduce computational complexity, which is important for real-time performance, especially on an embedded computer with limited resource.} to travel towards an object $x^{*}_k$ under the void probability functional constraint to rapidly minimize estimation uncertainty. We label the three possible UAV destinations under three control actions as points A, B, and C. For each possible trajectory, the path $u_{k+1:k+H}$ is evaluated 
	under the void probability constraint using \eqref{eq_void_prob} with a predefined probability bound $B_{\min} \in [0,1] $. Our trajectory planning algorithm selects the first control action that yields the trajectory that satisfies our void constraints in order of point A, B, and C. 
	
	However, we recognize that \hl{three traversal paths to point A, B, C of the object of interest $x_k^*$ may violate the void bound $B_{\min}$ due to the existence of other objects---see \eqref{eq_void_prob}}. Therefore we define a discrete set of actions $\mathbb{U}_k$ for the UAV to investigate in such an event \hl{(see Alg.~\ref{algo_guava}, Line 17)}. 
	The discrete space $\mathbb{U}_k$ consists of $|\mathbb{U}_k|$ actions that direct the UAV to change its heading to one of the following $[0, 2\pi/|\mathbb{U}_k|, \dots, 2\pi - 2\pi/|\mathbb{U}_k|]^T$ and then traverse along with the selected heading. As before, we 
	select the possible control action $u^c_k$ that directs the UAV towards the selected object $x_k^{*}$ such that:
	\vspace{-0.4cm}
	\begin{align} 
	u^c_k =& \argmin_{u^c \in \mathbb{U}_k} d(u^c,x^{*}_k) ,\\
	\text{subject to } &B_{\pi_k}(u_{k+1:k+H},r_{\min}) \geq B_{\min} \label{eq_find_uc}.
	\end{align}

	\noindent\textbf{Proposition 1.~}\label{prop_1} \hl{\textit{The selected control action $u^c_k$ from Alg.~\ref{algo_guava} always maintains the UAV at a desired safe distance $r_{\min}$ to any objects in $x_k$ with a lower bound probability of $B_{\min}$.}}
	
	\begin{proof}
		\hl{
			Let $u \in u_{k+1:k+H} $ be any position of the UAV generated from the control action $u_c^k$. According to the definition in \eqref{eq_void_prob}, the probability that $u$ maintains a safe distance to any objects in $\mathbb{X}_k$, $\mathrm{Pr}(u|x_k,r_{\min})$ is defined as:
			\begin{equation}
			\resizebox{.88\hsize}{!}{$ \mathrm{Pr}(u|x_k,r_{\min}) = \min_{\substack{ {x \in X_{k}} }} B_{\pi_k(x|\cdot)}\big(V(u, r_{\min})\big) \geq B_{\min} $}
			\end{equation}
			following the constraint in \eqref{eq_find_uc}. }
	\end{proof}
	
	\setlength{\textfloatsep}{0.1cm}
	\setlength{\floatsep}{0.1cm}
	
	\begin{algorithm}[!t]
		\captionof{algo}{\textit{LAVAPilot}} 
		\label{algo_guava}
		\begin{algorithmic}[1]
			\State $\mathbb{N}_t = \{1,\dots,N_t\}$ \Comment {Objects left to localize }
			\For{\textbf{each} $\lambda \in \mathbb{N}_t$}
			\If{$\sigma^{(x)} < \sigma_{\min}$} 
			\State $\mathbb{N}_t = \mathbb{N}_t \backslash \{\lambda\}$ \Comment{Remove found object $\lambda$ from list}
			\EndIf
			\EndFor
			\If{$\mathbb{N}_t \neq \emptyset$ and $k \mod H == 0$} 
			\State $u^c_k = \emptyset$ \Comment {Initialize the control action }    
			\State $x^{*}_k = \argmin\limits_{x \in \bigcup\limits_{\lambda \in \mathbb{N}_t} x_k} \sigma^{(x)}$ \Comment {Select lowest uncertainty} 
			\State $\mathbb{U}_{\text{toward}} = \{\text{A,~B,~C}\}$ as in Fig. \ref{fig_Move_toward_an_object_with_void} for $x^{*}_k$ with $r_{\min}$
			\For{\textbf{each} $u^c \in \mathbb{U}_{\text{toward}}$}
			\State Calculate $u_{k+1:k+H}$ using 1-step Euler method 
			\If{$B_{\pi}(u_{k+1:k+H},r_{\min}) \geq B_{\min}$}
			\State $u^c_k = u^c$; \textbf{break}
			\EndIf
			\EndFor
			\If {$u^c_k = \emptyset$} \Comment{Use discrete action space $\mathbb{U}_k$}
			\State $u^c_k = \argmin_{u^c \in \mathbb{U}_k} d(u^c,x_k^{*})$ 
			\State such that $B_{\pi_k}(u_{k+1:k+H},r_{\min}) \geq B_{\min}$      
			\EndIf
			\State \Return $u^c_k$
			\EndIf
			
		\end{algorithmic}
	\end{algorithm}

	\begin{figure*} [!t]
		\centering
		\includegraphics[width=0.9\textwidth]{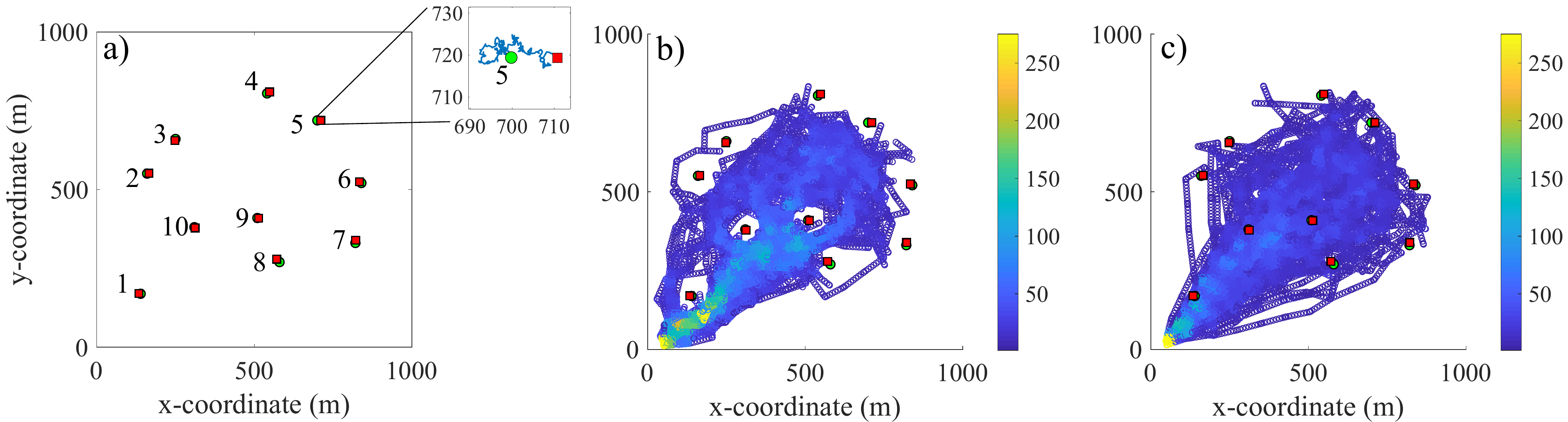}
		\vspace{-0.2cm}
		\caption{a) Ground truth for ten mobile objects wandering in the search area, \textit{inset}: a trajectory of a random walk object. b) UAV's trajectory heat map using the LAVAPilot control algorithm over 100 MC trials using the void constraint ($r_{\min} = 50$~m); c) Without using the void constraint ($r_{\min} = 0.001$~m). $\bigcirc/\Box$ denotes the start/stop positions of each object. We can clearly see the effectiveness of the void constraint in preventing a close approach by the aerial robot.}
		\vspace{-0.6cm}
		\label{fig:trajectory_heat_map}
	\end{figure*}
	
	\section{System Architecture and Implementation}
	
	Our autonomous robot system is illustrated in Fig. \ref{fig_uav_system}. The overall platform has a mass of 1280~g---a mass less than 2000~g is an attractive mass class for operating a UAV without a Remote Pilot License (RePL) in certain jurisdictions to \textit{i)}~ensure accessibility of the technology, \textit{ii)}~and portability of the drone platform. We summarize the main hardware components below: 
	\textit{i)} \textit{3DR Iris+}: an off-the-shelf, low-cost, and lightweight quad-copter UAV with limited battery power, turning rate, and velocity; \textit{ii)} \textit{2-element Yagi antenna}: a directional VHF antenna to receive transmitted signals from radio tags; \textit{iii)} \textit{HackRF One SDR}: a sensor system built with a software-defined radio (SDR) and the antenna element; and \textit{iv)} \textit{Raspberry Pi 3 Model A+ -- companion computer}: an embedded compute module with a quad-core ARM Cortex-A53 @ 1.4~GHz and a 512~MB SDRAM that executes algorithms to perform signal detection, planning and tracking. The sensor system (the directional antenna, SDR, signal detection algorithm) measures the received signal strength indicator (RSSI) of multiple radio-tags simultaneously.

	Notably, we used the physical components of the proposed sensor system reported in~\cite{hoa2017icra}. However, in contrast to \cite{hoa2017icra}, wherein an information-based tracking and control algorithm is executed on a ground control system in the loop, our proposed planning for tracking algorithm is performed \textit{directly} on the embedded computing module. The planning algorithm communicates directly with the ArduPilotMega (AMP) firmware through the MAVLink protocol. This enables a fully independent and autonomous aerial system. 

	Our proposed aerial robot design leads to enlarging the potential search area by \textit{i)}~removing the distance constraint imposed by the UAV and the GCS communication radio channel, and \textit{i)}~eliminating the power-intensive transmissions and receipts of messages to and from the GCS to conserve battery life. Most importantly, \textit{our design provides a highly portable solution for conservation fieldwork}.

	\section{Simulation Experiments}
	We validate our proposed algorithm (\textit{LAVAPilot}) through a simulated environment and compare it with information-based planning strategies in~\cite{hoa2017icra} and ~\cite{cliff2015online}. 
	
	\noindent\textbf{Experimental settings:} In our simulation, there are ten mobile radio-tagged objects walking randomly in a search area of $1000$~m $\times 1000$~m as depicted in Fig.~\ref{fig:trajectory_heat_map}a. The altitude of the UAV is maintained at $30$~m above ground level (AGL), while the heights of radio tags are fixed at $1$~m (AGL) to limit the planning scope to a 2D domain. 
	The maximum flight time is set to $3000$~s. The UAV is designed to maintain a safe distance $r_{\min} = 50$~m if the void constraint is applied and $r_{\min} = 0.001$~m if the void constraint is not applied. The void probability is lower bound $B_{\min} =0.8$, \ie, any control actions with the void probability smaller than $B_{\min}$ are discarded. The number of discrete actions we used is $|\mathbb{U}_k| = 12$, \ie, the allowed heading changes are $\{0,\pi/6,\dots,11\pi/6\}$. We set the look-a-head horizon time $H = 11$, measurement duration $T_0=1$~s, and the number of particles $N = 10,000$. Recall, an object is considered tracked and localized if its estimation uncertainty is smaller than $\sigma_{\min} = 35$~m.  
	
	\noindent\textbf{Performance evaluation metrics:} The algorithms are evaluated based on the following metrics:
	\begin{itemize}
		\item \textit{Root Mean Square (RMS) (m): } between the estimated object positions versus its ground truths. Let $\hat{X} = [\hat{x}_1,\dots,\hat{x}_{N_t}]^T \in \mathbb{X}$ be an estimate of its ground truth $X = [x_1,\dots,x_{N_t}]^T \in \mathbb{X}$. The RMS is the square root of the mean square error, given by:\break
		$\text{RMS} = \big(\sum\limits_{i=1}^{N_t}\big[d(\hat{x}_i,x_i)\big]^2\big/N_t)^{1/2}$.
		\item \textit{Standard deviations (Stdev) of estimated results:} via \eqref{eq:stdev_def}.
		\item \textit{Planning time (s):} time to compute the optimal control actions for the UAV subject to the void constraint.
		\item \textit{Flight time (s):} time for the UAV to search and localize all of the objects of interest. 
	\end{itemize}
	The reported results are averaged over 100 Monte-Carlo (MC) runs. Simulation experiments are performed on a PC using a single core of Intel Core Processor (Broadwell) @ 2.4 GHz and MATLAB-2019b. 

	\noindent\textbf{Performance evaluation with (w) void and without (w/o) void:} Fig.~\ref{fig:trajectory_heat_map}bc depicts the trajectory heat-map of the UAV using our proposed LAVAPilot control algorithm w void, and w/o void over 100 MC runs. The results demonstrate that LAVAPilot w void successfully maintains a safe distance from all of the objects of interest while  tracking and localizing these mobile objects. Further, as shown in Table~\ref{tab:sim_comparison_lava_with_others}, there is no significant performance difference whether the void constraint is applied or not. Hence, the proposed LAVAPilot strategy subject to a void constraint can maintain a safe distance to all of the objects while not compromising in terms of tracking and planning performance. 
	
	\noindent\textbf{Comparing LAVAPilot to Other Planning Methods:} We compare our proposed LAVAPilot algorithm with the information-based methods using R\'{e}nyi divergence \cite{hoa2017icra} and Shannon entropy~\cite{cliff2015online,hoa2018acra} under the same void constraint in \eqref{eq_find_uc}. Table~\ref{tab:sim_comparison_lava_with_others} lists average performance in terms of RMS, standard deviation (Stdev), planning time, and travel distance for each approach. The results demonstrate the task-based reward using LAVAPilot performs as well as the information-based methods under a void constraint in terms of localization accuracy (RMS values) but with over 98.5\% reduction in computational costs (compared to Shannon) albeit with a 17\% increase in flight time (compared to R\'{e}nyi) against the best performing information-based methods on each measure. Table~\ref{tab:sim_comparison_lava_with_others} also shows that applying the void constraint helps to reduce 17\% of planning time for information-based planning strategies since it discarded any control actions that violate the void constraint, thus leading to an improvement in planning time. It is expected that applying the void constraint results in the longer flight time for LAVAPilot since there are a few potentially useful trajectories to reduce the object's uncertainty that are not considered due to the void constraint.

	\begin{table}[!t]
		\centering
		\vspace{0.2cm}
		\caption{Comparing LAVAPilot over 100 Monte-Carlo runs for the same object tracking and localization task.}
		\label{tab:sim_comparison_lava_with_others}
		\resizebox{0.45\textwidth}{!}{%
			\begin{tabular}{|l|l|c|c|c|c|}
				\hline
				\textbf{Strategy} & \textbf{Constraint} & \multicolumn{1}{l|}{\textbf{\begin{tabular}[c]{@{}l@{}}RMS\\ (m)\end{tabular}}} & \multicolumn{1}{l|}{\textbf{\begin{tabular}[c]{@{}l@{}}Stdev\\ (m)\end{tabular}}} & \multicolumn{1}{l|}{\textbf{\begin{tabular}[c]{@{}l@{}}Planning\\ time (s)\end{tabular}}} & \multicolumn{1}{l|}{\textbf{\begin{tabular}[c]{@{}l@{}}Flight \\ time (s)\end{tabular}}} \\ \hline
				\textbf{LAVAPilot} & \multirow{3}{*}{w Void} & \textbf{18.4} & \textbf{14.4} & \textbf{0.06} & 193.3 \\ \cline{1-1} \cline{3-6} 
				\textbf{R\'{e}nyi} & & 19.7 & 15.8 & 4.33 & \textbf{165.3} \\ \cline{1-1} \cline{3-6} 
				\textbf{Shannon} & & 19.5 & 15.9 & 4.03 & 165.7 \\ \hline
				\textbf{LAVAPilot} & \multirow{3}{*}{w/o Void} & \textbf{18.6} & \textbf{14.8} & \textbf{0.04} & 183.5 \\ \cline{1-1} \cline{3-6} 
				\textbf{R\'{e}nyi} & & 21.7 & 16.0 & 5.20 & 160.1 \\ \cline{1-1} \cline{3-6} 
				\textbf{Shannon} & & 20.9 & 16.4 & 4.92 & \textbf{157.0} \\ \hline
			\end{tabular}%
		}
	\end{table}

\begin{figure} [!t]
		\centering
		\vspace{0.2cm}
		\includegraphics[width=0.3\textwidth]{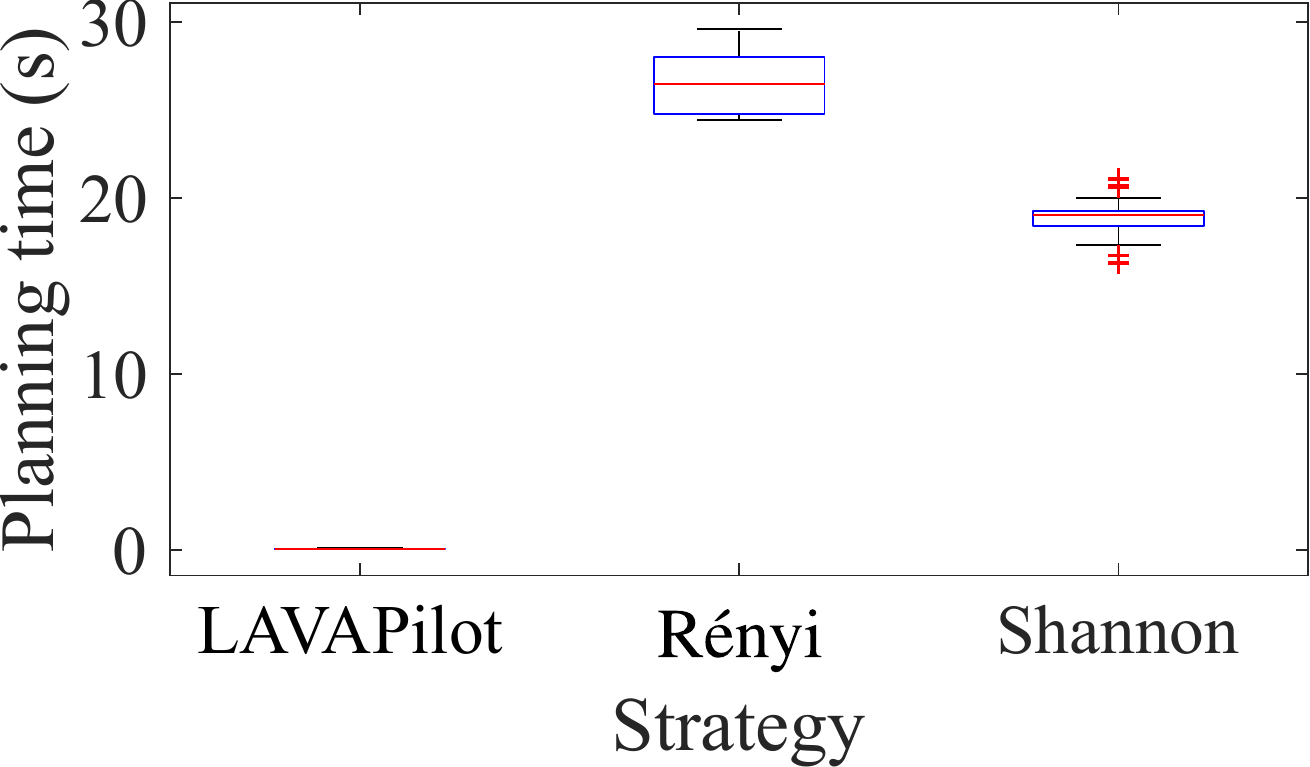}
		\vspace{-0.2cm}
		\caption{Planning time of different strategies on a Raspberry Pi 3 Model A+.}
		\label{fig:Pi_planning_time}
	\end{figure}

	\noindent\textbf{Computational investigations on an embedded computer:} The previous simulation experiments are conducted on a desktop PC, which cannot be carried by the UAV. In this experiment, we investigate the computational performance directly on an embedded computer (Raspberry Pi 3 Model A+) to validate our proposed approach (the same model employed in our aerial robot). All of the simulation settings are kept are the same, while the tracking and planning algorithm is implemented using Rust programming language instead of MATLAB on the Raspberry Pi 3 Model A+. 
	Fig.~\ref{fig:Pi_planning_time} and Table~\ref{tab:computation_cost_pi} present the performance comparison of planning time between LAVAPilot versus R\'{e}nyi and Shannon planning strategies. The results validate that our LAVAPilot control algorithm can perform in real-time environments on an embedded computer with a maximum planning time of $0.1$~s. In contrast, the information-based planning strategies (\ie, R\'{e}nyi and Shannon) require excessive computational power resulting in prohibitively high planning times, which prevents them from being implementing directly on a low power embedded compute module such as the Raspberry Pi.

	\begin{table}[!t]
		\centering
		\vspace{0.2cm}
		\caption{Performance comparison of planning time of different strategies for computing optimal control actions on a Raspberry Pi 3 Model A+.}
		\label{tab:computation_cost_pi}
		\resizebox{0.45\textwidth}{!}{%
			\begin{tabular}{|l|r|r|r|r|}
				\hline
				\textbf{Strategy} & \multicolumn{1}{c|}{\textit{\begin{tabular}[c]{@{}c@{}}Planning time\\ Average (s)\end{tabular}}} & \multicolumn{1}{c|}{\textit{\begin{tabular}[c]{@{}c@{}}Planning time\\ Min (s)\end{tabular}}} & \multicolumn{1}{c|}{\textit{\begin{tabular}[c]{@{}c@{}}Planning time\\ Max (s)\end{tabular}}} & \multicolumn{1}{c|}{\textit{\begin{tabular}[c]{@{}c@{}}Planning time\\ Median (s)\end{tabular}}} \\ \hline
				\textbf{LAVA} & \textbf{0.05} & \textbf{0.01} & \textbf{0.10} & \textbf{0.03} \\ \hline
				\textbf{R\'{e}yni} & 26.00 & 24.57 & 28.72 & 25.18 \\ \hline
				\textbf{Shannon} & 18.03 & 16.32 & 19.02 & 18.40 \\ \hline
			\end{tabular}%
		}
	\end{table}
	
	\section{Field Experiments}\label{sec:experiments}
	We conducted six autonomous missions and other experiments designed to test the components of our proposed system. Notably, our experiments were based on the guidelines from CASA (Civil Aviation Safety Authority, Australia) and our university concerning the conduct of UAV research. Given the requirement of operating in autonomous manners, our flight zone, as well as the experiment's scope, is confined within university-owned property allocated for UAV trials. 

	\subsection{Experimental settings}
	
	We consider an example scenario of localizing four stationary VHF radio tags with frequencies ranging from $150 - 152$ MHz in an area of $75~\text{m} \times 300$ m to demonstrate our system performance and capability. 
	In this field trial, our goal is to demonstrate the effectiveness of the void probability formulation using the minimum radius $r_{\min} = 15$~m compared to the control action w/o void (by setting $r_{\min} = 0.001$~m) with a fully autonomous aerial robot. Although our approach is formulated and implemented to track and locate mobile objects, 
	to ensure the reproducibility of both planning methods, we use stationary radio-tags instead of mobile tags. For the tracking task, we set $\sigma_{\min} = 15$~m to confirm the localization of an object.
	
\begin{figure} [!t]
		\centering
		\vspace{0.2cm}
		\includegraphics[clip,width=0.4\textwidth]{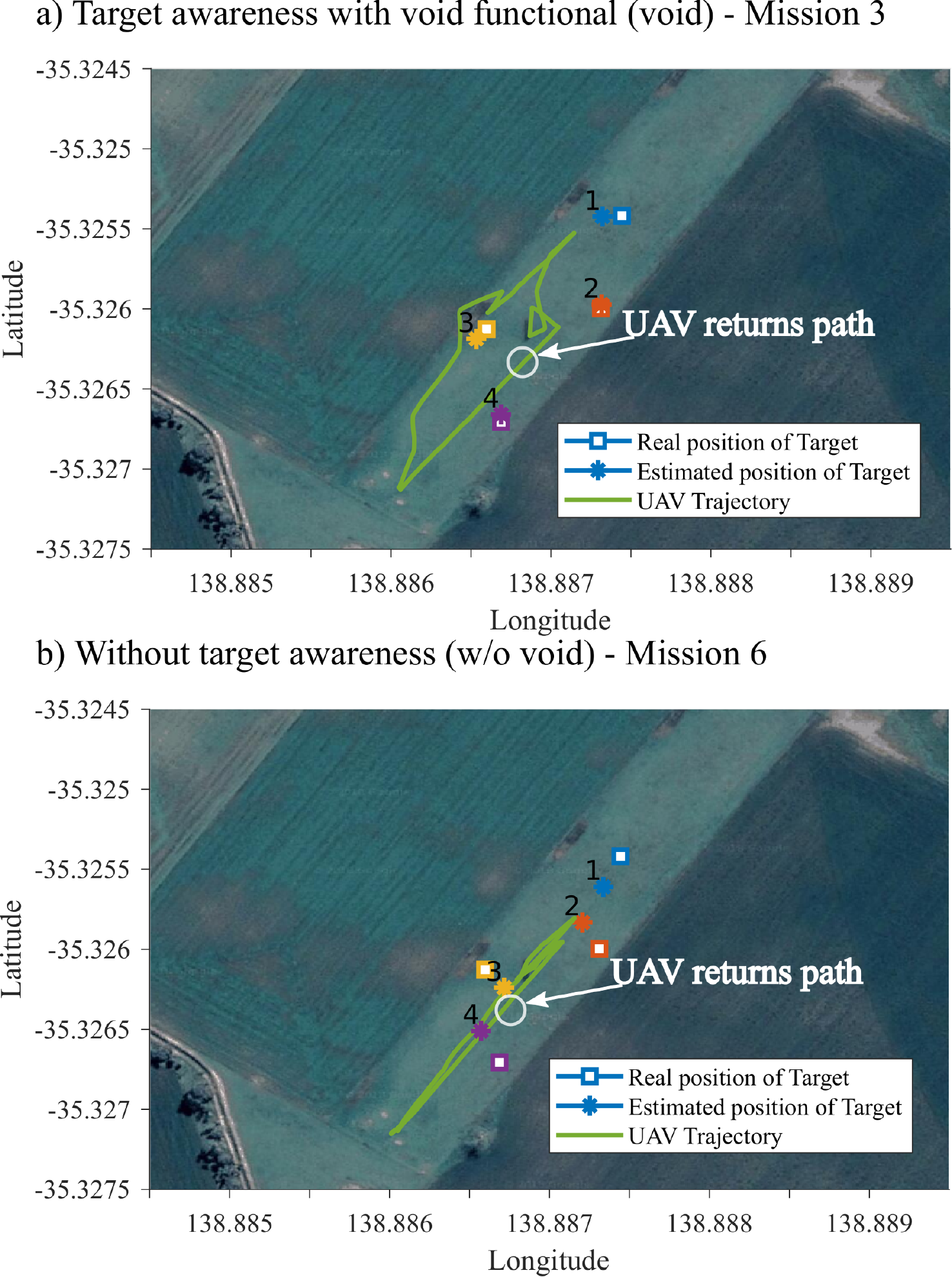}
		\vspace{-0.2cm}
		\caption{Field trial missions to localize four stationary tags. Stationary tags are used to ensure a fair comparison between w void and w/o void strategies. } 
		\label{fig_Field_Trial_Void_vs_NoVoid}
	\end{figure}

	We set the UAV to a fixed altitude of $30$ m (AGL) throughout the whole experiment and subsequently switch to 
	the tracking and planning algorithm onboard the UAV 
	that follows Alg.\ref{algo_guava}--LAVAPilot. After finishing the localization tasks, the UAV's altitude was increased to $80$~m (AGL) to minimize potential disturbances to the hypothetically localized radio-tagged wildlife before flying in a straight path back to its home position (as shown in Fig.~\ref{fig_Field_Trial_Void_vs_NoVoid}).
	We conduct six fully autonomous flights, including four flights w void ($r_{\min} = 15$ m) and two flights w/o void ($r_{\min} = 0.001$~m).  
	
	\begin{figure} [!t]
		\centering
		\vspace{0.2cm}
		\includegraphics[clip,width=0.45\textwidth]{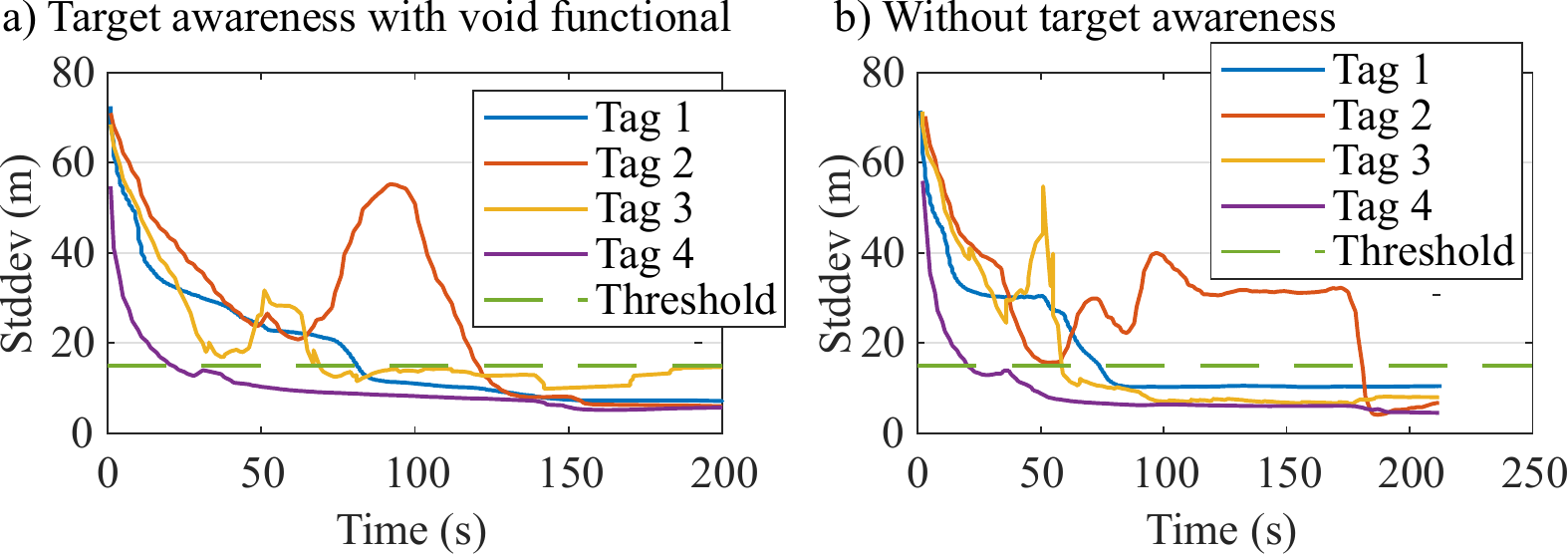}
		\vspace{-0.2cm}
		\caption{The estimation uncertainty $\sigma^{(x)}$ for each radio-tag object for the mission in Fig. \ref{fig_Field_Trial_Void_vs_NoVoid}. We can see LAVApilot shifting to the next object as each object is localized (meeting the threshold). Notably, even after localization, we continue to track all the visible objects.}
		\label{fig_Field_Trial_Void_vs_NoVoid_Std}
	\end{figure}
	
	\subsection{Results}
	Fig.~\ref{fig_Field_Trial_Void_vs_NoVoid} depicts the UAV trajectories in two different scenarios: localize objects w void (Fig.~\ref{fig_Field_Trial_Void_vs_NoVoid}a) and w/o void (Fig.~\ref{fig_Field_Trial_Void_vs_NoVoid}b). We can see that the UAV trajectory (green path) w void navigates the UAV around the radio-tags of interest. In contrast, results in trajectory planning w/o void plot a path through the object distributions to track and localize the objects without maintaining a safe distance. The field experiments confirm the simulation results---our proposed trajectory planning algorithm for tracking prevents the UAV approach to the objects of interest whilst achieving its objective autonomously. 
	
	Fig.~\ref{fig_Field_Trial_Void_vs_NoVoid_Std} depicts the estimated uncertainty for each object over time for both planning methods. The results show that there are differences in the estimated uncertainty among radio-tags; however, over time, the planning strategy effectively maneuvers the UAV to localize the radio-tags. In particular, radio-tag $2$ shows the highest uncertainty fluctuations over time. One explanation for this case is that, in practice, the transmitted power of each radio tag is different while we assumed the same transmitted power for each radio-tag. This added uncertainty causes apparent estimation errors. 
	
	Table~\ref{table_localization_void_vs_novoid} presents a quantitative summary of the localization results in terms of Root Mean Square (RMS) and total flight times. The results show that our tracking algorithm can localize four radio-tags accurately, and there is no significant \hl{localization} accuracy difference between the two investigated planning methods. Hence, the results confirm the effectiveness of our proposed algorithm w void where the UAV is still able to effectively and accurately localized all objects of interest without approaching them.

	\begin{table}[!tb]
		\centering
		\vspace{0.2cm}
		\caption{Localization performance over six field missions to locate four VHF radio-tags with and w/o void awareness.}
		\label{table_localization_void_vs_novoid}
		\resizebox{0.45\textwidth}{!}{%
			\begin{tabular}{ll|r|r|r|r|r|r|}
				\cline{3-8}
				& & \multicolumn{5}{c|}{\textbf{RMS (m)}} & \multicolumn{1}{l|}{\multirow{2}{*}{\textbf{\begin{tabular}[c]{@{}l@{}}Flight\\ Time (s)\end{tabular}}}} \\ \cline{3-7}
				& & \multicolumn{1}{l|}{\textbf{Tag 1}} & \multicolumn{1}{l|}{\textbf{Tag 2}} & \multicolumn{1}{l|}{\textbf{Tag 3}} & \multicolumn{1}{l|}{\textbf{Tag 4}} & \multicolumn{1}{l|}{\textbf{Mean}} & \multicolumn{1}{l|}{} \\ \hline
				\multicolumn{1}{|l|}{\textbf{Mission 1}} & \multirow{4}{*}{Void} & 31.4 & 24.5 & 5.1 & 19.5 & 20.1 & 189 \\ \cline{1-1} \cline{3-8} 
				\multicolumn{1}{|l|}{\textbf{Mission 2}} & & 41.0 & 35.0 & 18.6 & 25.7 & 30.1 & 150 \\ \cline{1-1} \cline{3-8} 
				\multicolumn{1}{|l|}{\textbf{Mission 3}} & & 11.1 & 2.7 & 8.9 & 5.2 & 7.0 & 200 \\ \cline{1-1} \cline{3-8} 
				\multicolumn{1}{|l|}{\textbf{Mission 4}} & & 14.3 & 10.8 & 3.4 & 6.4 & 8.7 & 176 \\ \hline
				\multicolumn{1}{|l|}{\textbf{Mission 5}} & \multirow{2}{*}{\begin{tabular}[c]{@{}l@{}}w/o\\ Void\end{tabular}} & 21.4 & 15.0 & 16.2 & 10.0 & 15.7 & 164 \\ \cline{1-1} \cline{3-8} 
				\multicolumn{1}{|l|}{\textbf{Mission 6}} & & 23.3 & 20.6 & 16.5 & 24.1 & 21.1 & 213 \\ \hline
			\end{tabular}%
		}
	\end{table}

	\section{Conclusions} \label{sec_conclusion}
	We have demonstrated the first \textit{fully autonomous} aerial system capable of real-time planning for tracking with situational awareness to maintain a safe distance from the UAV to the objects of interest. We formulated a new task-based planning method incorporating a void probability functional with low computational demand. Our field experiments confirm the effectiveness of our proposed autonomous aerial robot system for the tracking task. Future work can focus on using other filtering algorithms (e.g., Box-particle filter~\cite{merlinge2019box}) to improve tracking accuracy and computational time.
	\section{ACKNOWLEDGMENT}
	This work was supported by the Australian Research Council (LP160101177), Department of Parks and Wildlife in Western Australia, the Shultz Foundation, and the Defense Science and Technology Group.


\begin{thebibliography}{10}
\providecommand{\url}[1]{#1}
\csname url@rmstyle\endcsname
\providecommand{\newblock}{\relax}
\providecommand{\bibinfo}[2]{#2}
\providecommand\BIBentrySTDinterwordspacing{\spaceskip=0pt\relax}
\providecommand\BIBentryALTinterwordstretchfactor{4}
\providecommand\BIBentryALTinterwordspacing{\spaceskip=\fontdimen2\font plus
\BIBentryALTinterwordstretchfactor\fontdimen3\font minus
  \fontdimen4\font\relax}
\providecommand\BIBforeignlanguage[2]{{%
\expandafter\ifx\csname l@#1\endcsname\relax
\typeout{** WARNING: IEEEtran.bst: No hyphenation pattern has been}%
\typeout{** loaded for the language `#1'. Using the pattern for}%
\typeout{** the default language instead.}%
\else
\language=\csname l@#1\endcsname
\fi
#2}}

\bibitem{Murphy2008}
R.~R. Murphy, S.~Tadokoro, D.~Nardi, A.~Jacoff, P.~Fiorini, H.~Choset, and
  A.~M. Erkmen, ``Search and rescue robotics,'' in \emph{Springer handbook of
  robotics}.\hskip 1em plus 0.5em minus 0.4em\relax Springer, 2008, pp.
  1151--1173.

\bibitem{dames2017detecting}
P.~Dames, P.~Tokekar, and V.~Kumar,
  ``\href{http://journals.sagepub.com/doi/abs/10.1177/0278364917709507}{Detecting,
  localizing, and tracking an unknown number of moving targets using a team of
  mobile robots},'' \emph{The Int.J. of Robotics Research}, vol.~36, no. 13-14,
  2017.

\bibitem{chung2018a}
S.~{Chung}, A.~A. {Paranjape}, P.~{Dames}, S.~{Shen}, and V.~{Kumar}, ``A
  survey on aerial swarm robotics,'' \emph{IEEE Trans. on Robotics}, vol.~34,
  no.~4, pp. 837--855, Aug 2018.

\bibitem{wich2015drones}
S.~A. Wich, ``\href{http://drones.newamerica.org/primer/Chapter%207.pdf}{Drones
  and conservation},'' \emph{Drones and Aerial Observation: New Technologies
  for Property Rights, Human Rights, and Global Development}, 2015.

\bibitem{Shafer2019}
M.~W. Shafer, G.~Vega, K.~Rothfus, and P.~Flikkema, ``{UAV} wildlife
  radiotelemetry: System and methods of localization,'' \emph{Methods in
  Ecology and Evolution}, vol.~10, no.~10, pp. 1783--1795.

\bibitem{wikelski2007going}
M.~Wikelski, R.~W. s, N.~J. Kasdin, K.~Thorup, J.~A. Smith, and G.~W. Swenson,
  ``\href{http://jeb.biologists.org/content/210/2/181.short}{Going wild: what a
  global small-animal tracking system could do for experimental biologists},''
  \emph{J. of Exp. Biology}, vol. 210, no.~2, pp. 181--186, 2007.

\bibitem{selby2011autonomous}
W.~Selby, P.~Corke, and D.~Rus,
  ``\href{http://www.araa.asn.au/acra/acra2011/papers/pap165.pdf}{Autonomous
  aerial navigation and tracking of marine animals},'' in \emph{Proc. of the
  ACRA}, 2011.

\bibitem{zhou2013thermal}
D.~Zhou, ``\href{https://conservancy.umn.edu/handle/11299/144870}{Thermal
  image-based deer detection to reduce accidents due to deer-vehicle
  collisions},'' 2013.

\bibitem{olivares2015towards}
M.~A. Olivares-Mendez, C.~Fu, P.~Ludivig, T.~F. Bissyand{\'e}, S.~Kannan,
  M.~Zurad, A.~Annaiyan, H.~Voos, and P.~Campoy,
  ``\href{http://www.mdpi.com/1424-8220/15/12/29861/html}{Towards an autonomous
  vision-based unmanned aerial system against wildlife poachers},''
  \emph{Sensors}, vol.~15, no.~12, pp. 31\,362--31\,391, 2015.

\bibitem{gonzalez2016unmanned}
L.~F. Gonzalez, G.~A. Montes, E.~Puig, S.~Johnson, K.~Mengersen, and K.~J.
  Gaston, ``\href{http://www.mdpi.com/1424-8220/16/1/97/htm}{Unmanned Aerial
  Vehicles (UAVs) and artificial intelligence revolutionizing wildlife
  monitoring and conservation},'' \emph{Sensors}, vol.~16, no.~1, p.~97, 2016.

\bibitem{kays2011tracking}
R.~Kays, S.~Tilak, M.~Crofoot, T.~Fountain, D.~Obando, A.~Ortega, F.~Kuemmeth,
  J.~Mandel, G.~Swenson, T.~Lambert, \emph{et~al.},
  ``\href{http://crofoot.ucdavis.edu/uploads/5/2/9/1/52914015/kays_etal_2011_arts.pdf}{Tracking
  animal location and activity with an automated radio telemetry system in a
  tropical rainforest},'' \emph{The Computer Journal}, pp. 1931--1948, 2011.

\bibitem{thomas2012wildlife}
B.~Thomas, J.~D. Holland, and E.~O. Minot,
  ``\href{https://www.researchgate.net/publication/257816464_Wildlife_tracking_technology_options_and_cost_considerations}{Wildlife
  tracking technology options and cost considerations},'' \emph{Wildlife
  Research}, vol.~38, no.~8, pp. 653--663, 2012.

\bibitem{tremblay2017low}
J.~A. Tremblay, A.~Desrochers, Y.~Aubry, P.~Pace, and D.~M. Bird,
  ``\href{http://www.nrcresearchpress.com/doi/abs/10.1139/juvs-2016-0021}{A
  low-cost technique for radio-tracking wildlife using a small standard
  unmanned aerial vehicle},'' \emph{J. of Unmanned Veh. Syst.}, vol.~5, 2017.

\bibitem{webber2017radio}
D.~Webber, N.~Hui, R.~Kastner, and C.~Schurgers,
  ``\href{https://ieeexplore.ieee.org/abstract/document/7876260/}{Radio
  receiver design for Unmanned Aerial wildlife tracking},'' in \emph{Int. Conf.
  on Comp., Net. and Comm.}, 2017, pp. 942--946.

\bibitem{hodgson2016best}
J.~C. Hodgson and L.~P. Koh,
  ``\href{https://www.sciencedirect.com/science/article/pii/S0960982216303189}{Best
  practice for minimising unmanned aerial vehicle disturbance to wildlife in
  biological field research},'' \emph{Current Biology}, vol.~26, no.~10, pp.
  R404--R405, 2016.

\bibitem{mulero2017unmanned}
M.~Mulero-P{\'a}zm{\'a}ny, S.~Jenni-Eiermann, N.~Strebel, T.~Sattler, J.~J.
  Negro, and Z.~Tablado,
  ``\href{http://journals.plos.org/plosone/article/file?id=10.1371/journal.pone.0178448&type=printable}{Unmanned
  aircraft systems as a new source of disturbance for wildlife: A systematic
  review},'' \emph{PloS one}, vol.~12, no.~6, p. e0178448, 2017.

\bibitem{paranjape2018robotic}
A.~A. Paranjape, S.-J. Chung, K.~Kim, and D.~H. Shim, ``Robotic herding of a
  flock of birds using an unmanned aerial vehicle,'' \emph{IEEE Trans. on
  Robotics}, vol.~34, no.~4, pp. 901--915, 2018.

\bibitem{tokekar2010robotic}
P.~Tokekar, D.~Bhadauria, A.~Studenski, and V.~Isler,
  ``\href{http://onlinelibrary.wiley.com/doi/10.1002/rob.20364/full}{A robotic
  system for monitoring carp in Minnesota lakes},'' \emph{J. of Field
  Robotics}, vol.~27, no.~6, pp. 779--789, 2010.

\bibitem{huntsberger2011intel}
T.~{Huntsberger} and G.~{Woodward}, ``Intelligent autonomy for unmanned surface
  and underwater vehicles,'' in \emph{OCEANS'11 MTS/IEEE KONA}, Sep. 2011, pp.
  1--10.

\bibitem{tokekar2013tracking}
P.~{Tokekar}, E.~{Branson}, J.~{Vander Hook}, and V.~{Isler}, ``Tracking
  aquatic invaders: Autonomous robots for monitoring invasive fish,''
  \emph{IEEE Robotics Aut. Mag.}, vol.~20, no.~3, pp. 33--41, Sep. 2013.

\bibitem{seto2013intro}
M.~L. Seto, L.~Paull, and S.~Saeedi, \emph{Introduction to Autonomy for Marine
  Robots}.\hskip 1em plus 0.5em minus 0.4em\relax New York, NY: Springer New
  York, 2013, pp. 1--46.

\bibitem{vander2014cautious}
J.~Vander~Hook, P.~Tokekar, and V.~Isler,
  ``\href{http://onlinelibrary.wiley.com/doi/10.1002/rob.21499/abstract}{Cautious
  Greedy Strategy for Bearing-only Active Localization: Analysis and Field
  Experiments},'' \emph{J. of Field Robotics}, vol.~31, no.~2, pp. 296--318,
  2014.

\bibitem{hook2015algo}
J.~{Vander Hook}, P.~{Tokekar}, and V.~{Isler}, ``Algorithms for cooperative
  active localization of static targets with mobile bearing sensors under
  communication constraints,'' \emph{IEEE Trans. on Robotics}, vol.~31, no.~4,
  pp. 864--876, Aug 2015.

\bibitem{das2015data}
J.~Das, F.~Py, J.~B. Harvey, J.~P. Ryan, A.~Gellene, R.~Graham, D.~A. Caron,
  K.~Rajan, and G.~S. Sukhatme, ``Data-driven robotic sampling for marine
  ecosystem monitoring,'' \emph{The Int.J. of Robotics Research}, vol.~34,
  no.~12, pp. 1435--1452, 2015.

\bibitem{frew2005radio}
E.~Frew, C.~Dixon, B.~Argrow, and T.~Brown, ``Radio source localization by a
  cooperating uav team,'' in \emph{Infotech@ Aerospace}, 2005.

\bibitem{stachura2011cooperative}
M.~Stachura and E.~W. Frew, ``Cooperative target localization with a
  communication-aware unmanned aircraft system,'' \emph{J. of Guidance,
  Control, and Dynamics}, vol.~34, no.~5, pp. 1352--1362, 2011.

\bibitem{jensen2014monte}
A.~M. Jensen, D.~K. Geller, and Y.~Chen,
  ``\href{https://link.springer.com/article/10.1007/s10846-013-9949-9}{Monte
  Carlo simulation analysis of tagged fish radio tracking performance by
  swarming unmanned aerial vehicles in fractional order potential fields},''
  \emph{J. of Intel. \& Robotic Sys.}, vol.~74, no. 1-2, pp. 287--307, 2014.

\bibitem{dos2014small}
G.~A.~M. Dos~Santos, Z.~Barnes, E.~Lo, B.~Ritoper, L.~Nishizaki, X.~Tejeda,
  A.~Ke, H.~Lin, C.~Schurgers, A.~Lin, \emph{et~al.},
  ``\href{https://ieeexplore.ieee.org/abstract/document/7035779/}{Small
  unmanned aerial vehicle system for wildlife radio collar tracking},'' in
  \emph{IEEE Proc. of MASS}, 2014, pp. 761--766.

\bibitem{vonehr2016software}
K.~VonEhr, S.~Hilaski, B.~E. Dunne, and J.~Ward,
  ``\href{https://ieeexplore.ieee.org/abstract/document/7535285/}{Software
  Defined Radio for direction-finding in UAV wildlife tracking},'' in
  \emph{IEEE Int. Conf. on Electro Inf. Tech.}, 2016, pp. 0464--0469.

\bibitem{cliff2015online}
O.~M. Cliff, R.~Fitch, S.~Sukkarieh, D.~Saunders, and R.~Heinsohn,
  ``\href{http://www.roboticsproceedings.org/rss11/p42.pdf}{Online Localization
  of Radio-Tagged Wildlife with an Autonomous Aerial Robot System},'' in
  \emph{Robotics: Science and Systems}, 2015.

\bibitem{hoa2017icra}
H.~V. Nguyen, M.~Chesser, L.~P. Koh, S.~H. Rezatofighi, and D.~C. Ranasinghe,
  ``Trackerbots: Autonomous unmanned aerial vehicle for real-time localization
  and tracking of multiple radio-tagged animals,'' \emph{J. of Field Robotics},
  vol.~36, no.~3, pp. 617--635, 2019.

\bibitem{bayram2017tracking}
H.~Bayram, N.~Stefas, K.~S. Engin, and V.~Isler,
  ``\href{https://ieeexplore.ieee.org/abstract/document/8250937/}{Tracking
  wildlife with multiple UAVs: System design, safety and field experiments},''
  in \emph{Proc. of MRS}, 2017, pp. 97--103.

\bibitem{hoa2019tbd}
H.~V. Nguyen, S.~H. Rezatofighi, B.-N. Vo, and D.~C. Ranasinghe, ``Online {UAV}
  path planning for joint detection and tracking of multiple radio-tagged
  objects,'' \emph{IEEE Trans. on Signal Proc.}, vol.~67, no.~20, pp.
  5365--5379, 2019.

\bibitem{Gordon1993}
N.~J. Gordon, D.~J. Salmond, and A.~F. Smith,
  ``\href{http://ieeexplore.ieee.org/stamp/stamp.jsp?arnumber=210672}{Novel
  approach to nonlinear/non-Gaussian Bayesian state estimation},'' \emph{IEE
  Proc. F - Radar and Signal Proc.}, vol. 140, no.~2, pp. 107--113, 1993.

\bibitem{daum2005nonlinear}
F.~Daum, ``Nonlinear filters: beyond the kalman filter,'' \emph{IEEE Aero. and
  Elect. Syst. Mag.}, vol.~20, no.~8, pp. 57--69, 2005.

\bibitem{wc1974microwave}
W.~Jakes,
  \emph{\href{https://courses.engr.illinois.edu/ece458/Jakes_Book_optim.pdf}{Microwave
  mobile communications}}.\hskip 1em plus 0.5em minus 0.4em\relax Wiley, NY,
  1974.

\bibitem{beard2017void}
M.~A. Beard, B.-T. Vo, B.~N. Vo, and S.~Arulampalam,
  ``\href{http://ieeexplore.ieee.org/stamp/stamp.jsp?arnumber=7967827}{Void
  probabilities and cauchy-schwarz divergence for generalized labeled
  multi-bernoulli models},'' \emph{IEEE Trans. on Signal Proc.}, vol.~65, 2017.

\bibitem{pukelsheim2006optimal}
F.~Pukelsheim, \emph{Optimal design of experiments}.\hskip 1em plus 0.5em minus
  0.4em\relax SIAM, 2006.

\bibitem{hoa2018acra}
H.~V. Nguyen, H.~Rezatofighi, D.~Taggart, B.~Ostendorf, and D.~Ranasinghe,
  ``Trackerbots: Software in the loop study of quad-copter robots for locating
  radio-tags in a {3D} space,'' in \emph{Proc. of the ACRA}, 2018.

\bibitem{merlinge2019box}
N.~Merlinge, K.~Dahia, H.~Piet-Lahanier, J.~Brusey, and N.~Horri, ``A box
  regularized particle filter for state estimation with severely ambiguous and
  non-linear measurements,'' \emph{Automatica}, vol. 104, pp. 102--110, 2019.

\end{thebibliography}
\end{document}